# AiSearch: Interactive Multi-Modal Search with VLMs

Ali Koksal, Mei Chee Leong, Vicky Sintunata, Ching Ling Chin, and Wee Teck Fong

Institute of Advanced Intelligence and Computing (IAIC),
Agency for Science, Technology and Research (A*STAR), Singapore
Koksal_Ali@a-star.edu.sg

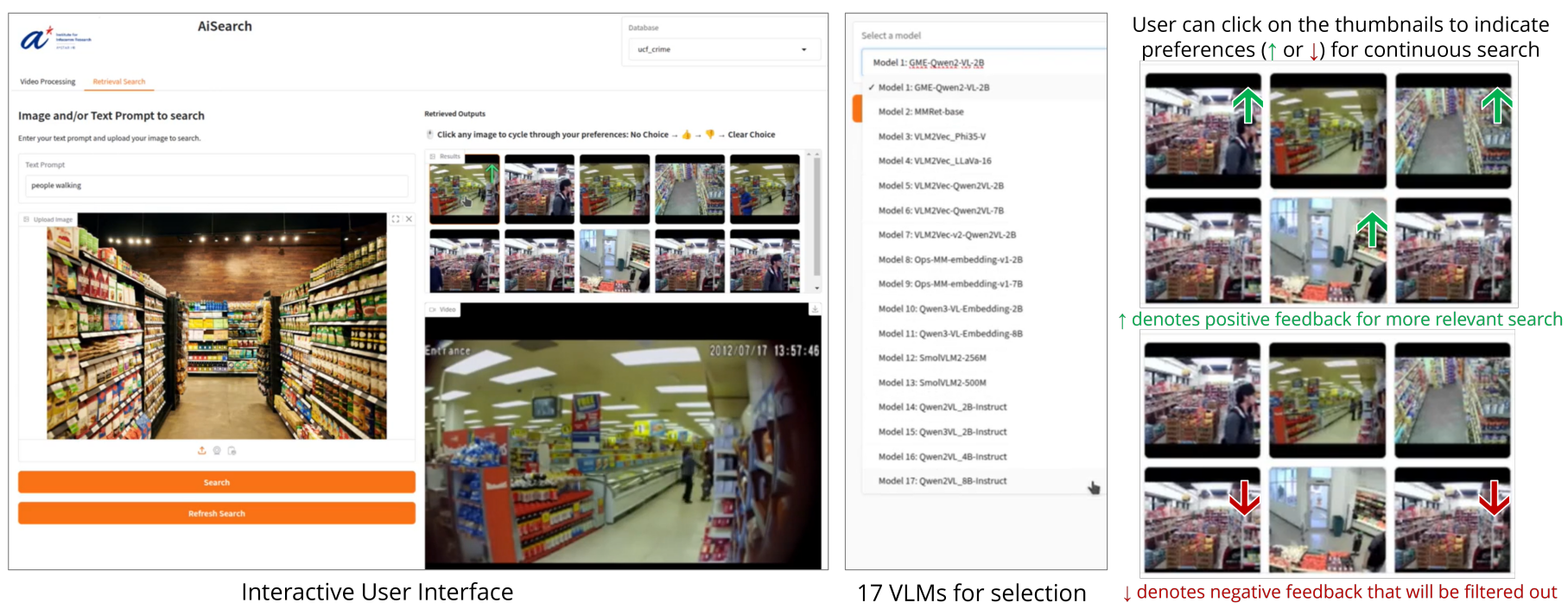

**Fig. 1:** AiSearch engine for interactive multi-modal search and refinement.

Modern retrieval systems must both be automated and interactive, allowing users to search and refine results in real-time. We present AiSearch, a flexible multi-modal retrieval framework that leverages the zero-shot capabilities of Vision-Language Models (VLMs) for natural language search over images and videos. AiSearch supports interactive search refinement through user feedback to tailor results to the user's intent, and allows visual benchmarking across multiple VLMs, enabling users to select the most suitable model for their task. The effectiveness the zero-shot retrieval approach is validated on the UCF-Crime dataset.

**System Design.** AiSearch is a modular AI engine that supports image, text, and hybrid queries for searching over large-scale image and video collections. The system consists of two processes (Figure 1 in Sup. Mat.). During Ingestion, videos and images are processed to extract embeddings using pre-trained VLMs, which are then indexed in a FAISS database for efficient retrieval. In the Search process, a user submits a query, and the system encodes the query to retrieve the top-$k$ most relevant results. AiSearch has integrated 17 VLMs, including both instruct and embedding models. Its modular architecture allows scalability with easy adoption of latest VLMs.

**Interactive Features.** AiSearch provides two key interactive capabilities. First, it enables continuous search with interaction. Users can iteratively refine search results through thumbs-up/down feedback, creating a continuous retrieval loop that progressively aligns results with user intent. Second, it facilitates visual benchmarking across VLMs. Users can visually benchmark and switch among different VLMs to compare retrieval performance and select the most effective model for their specific domain.

**Evaluation.** To validate our retrieval engine, we benchmarked multiple VLMs on the UCF-Crime dataset. AiSearch achieves performance up to 81.54% AUC using Ops-MM-embedding-v1-7B on anomaly detection task, without requiring explicit training. Additional implementation details, UI interactions, ablation studies, and comprehensive results are provided in the Sup. Mat. and the accompanying demo video.

This study is supported by the National Research Foundation, Singapore, under the Public Sector Translational R&D Grant Funding Initiative (TRANS Grant) (Grant No. TRANS2023-TGC05).

# Supplementary Material for AiSearch: Interactive Multi-Modal Search with VLMs

Ali Koksal, Mei Chee Leong, Vicky Sintunata, Ching Ling Chin, and Wee Teck Fong

Institute of Advanced Intelligence and Computing (IAIC),
Agency for Science, Technology and Research (A*STAR), Singapore
Koksal_Ali@a-star.edu.sg

## 1 Motivation

While automated video analysis tools are essential for managing vast amounts of multimedia, they often lack the nuanced understanding required for complex, open-ended exploratory tasks. A purely automated, static approach is rigid and cannot adapt to the evolving, subjective criteria of a user, whether investigating surveillance footage for anomalies, searching archives for specific events, or analyzing media for nuanced visual cues.

To bridge this gap, we propose AiSearch, a modular multi-modal retrieval framework that offers high adaptability and flexibility to fit diverse use cases across images and videos. Designed as a versatile AI engine, AiSearch allows users to formulate queries using natural language, images, or a combination of both. This multi-modal capability provides users with the freedom to express complex search intent without being restricted. Furthermore, the system incorporates user relevance feedback directly into the retrieval process. This enables users to actively guide the search, shifting from a single static query to an iterative loop that dynamically refines the results to match their specific intent.

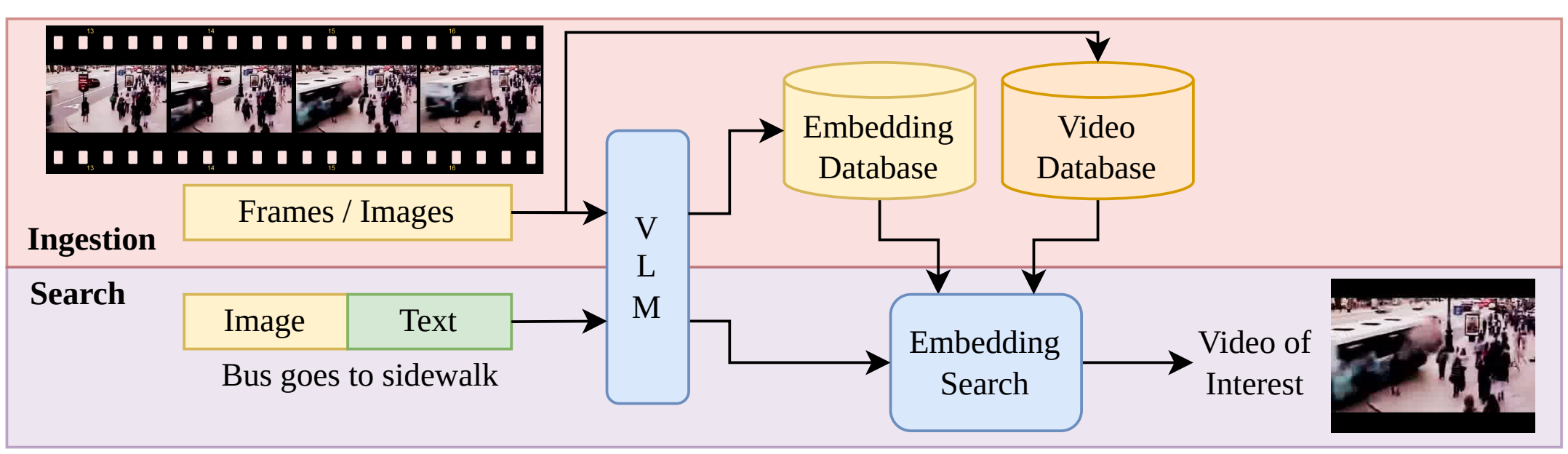

**Fig. 1:** AiSearch system architecture, illustrating the two main processes: Ingestion and Search. The Ingestion process builds searchable databases from source videos and images, while the Search process handles multi-modal user queries.

## 2 System Design

AiSearch is a modular AI engine designed for multi-modal search and retrieval, supporting queries in the form of images, text, or a combination of both. By enabling real-time user interaction, it facilitates efficient and scalable search over a large collection of videos. The framework supports VLMs integration, including both instruct models (e.g., SmolVLM2-500M [5], Qwen2-VL-2B-Instruct [11]), which are trained to generate natural language responses, and embedding models (e.g., VLM2Vec models [3], Qwen3-VL-Embedding-2B [4]), which are trained or finetuned by contrastive learning to map multi-modal inputs into high-dimensional embedding spaces. AiSearch's modular architecture allows easy adoption of the latest VLMs into the framework. Figure 1 illustrates the overall architecture, which consists of two independent processes: Ingestion and Search.

### 2.1 Ingestion Process

AiSearch constructs searchable databases from the target video collection. Raw videos are first processed by extracting frames and sampled uniformly, followed by computing embeddings using a pre-trained VLM. Extracted frames and embeddings are stored in databases, correspondingly. In addition to video retrieval, AiSearch supports image retrieval, where the images are processed similarly.

The input images (or video frames) are first processed by a visual encoder (e.g., CNN or Vision Transformer) to extract visual features, while text inputs are tokenized and encoded into token embeddings. These representations are then jointly processed through transformer layers to model cross-modal relationships. The embeddings from the activation of the final hidden layer captured high-level semantic representations, are extracted and pooled to obtain a fixed-size embedding for downstream tasks.

While earlier models such as CLIP [9] compute embeddings of text and images independently, the VLMs employed in AiSearch leverage textual prompts alongside visual inputs, enabling context-aware and task-adaptive representation learning. Recognizing that the performance of a VLM can vary by domain, AiSearch includes a visual benchmarking feature. It allows users to build the databases with various VLMs and execute the same query across multiple VLMs and visually compare the results, enabling users to identify most effective model for their specific needs.

In this work, we use a generic text prompt inspired by [3] *"Represent the given image with the following question: What is in the image?"* for embedding extraction. This prompt is designed to resemble a visual question answering task, encouraging the model to generate embeddings that reflect a general understanding of the scene depicted in each frame. As a result, the embeddings produced are not merely visual descriptors but are enriched with semantic information derived from the prompt, making them more effective for retrieval tasks. This prompt can be updated according to different use cases.

### 2.2 Search Process

The embedding of the query, whether text, image, or both, is computed by the same VLM used during the Ingestion process. AiSearch then computes similarity scores between the query embedding and the stored embeddings in the database. Frames (or images) are ranked based on cosine similarity, and the top-$k$ results are retrieved accordingly. To ensure scalable retrieval over large-scale video collections, we employ FAISS [2], which enables fast searching and sorting via cosine similarity.

After an initial retrieval, users often have a clearer idea of what they are looking for. AiSearch facilitates this refinement process via a thumbs-up/down feedback mechanism. For each retrieved item, users can provide explicit positive (**thumbs-up**) or negative (**thumbs-down**) feedback. This explicit feedback is leveraged to dynamically refresh and refine the search results. To process positive and negative feedback, we have distinct mechanisms:

**- Positive Feedback (Thumbs-Up):** When users thumbs-up specific results, new embeddings are computed by combining the original text prompt (if one was provided) with each of the selected, thumbed-up results. The aggregated mean of newly computed embeddings serves as the updated, highly-contextualized query embedding to retrieve a fresh set of results from the database.

**- Negative Feedback (Thumbs-Down):** When users thumbs-down results, those selected frames are filtered out from the retrieved list. However, to effectively guide the search away from undesired content, AiSearch also filters out conceptually similar items. It computes the cosine similarity between the embeddings of the thumbed-down results and the remaining retrieved results. Any results with a similarity score above a threshold (set to 0.95 by default) are automatically removed as well. Finally, new results from the database are appended to fill the refreshed list of retrieved results.

This feedback mechanism creates a powerful, continuous search loop where users and AiSearch collaborate to progressively tailor the search results to the user's specific intent. The accompanying demo video showcases this interactive workflow.

## 3 Benchmarking

Although AiSearch is a highly versatile framework capable of performing retrieval across any collection of visual media (e.g., single clips, extensive video archives, short videos, or image datasets) and across diverse domains - such as media broadcasting, sports analytics, or industrial inspection, we chose to rigorously validate its capabilities on a well-established research area. Specifically, we conducted a comprehensive benchmarking of our framework to evaluate its effectiveness in identifying anomalous events in CCTV surveillance videos.

**Dataset.** UCF-Crime [10] dataset is a large-scale benchmark for video anomaly detection, comprising long, untrimmed surveillance footage across 13 categories of anomalous events (e.g., robbery, assault, burglary). For evaluation, we used the standard test split, which includes 290 videos, 140 abnormal and 150 normal. These videos reflect diverse real-world scenarios with varying backgrounds, lighting conditions, and camera angles, making the detection task particularly challenging.

**Evaluation Metric.** Following existing works in video anomaly detection [10,12], we use the Area Under the ROC Curve (AUC) as the primary evaluation metric. AUC is computed using frame-level annotations of the test videos, providing a fine-grained assessment of the model's ability to distinguish between normal and abnormal frames.

**Implementation Details.** To estimate anomaly scores, we utilize a diverse pool of semantically related keywords. The abnormal pool consists of 84 keywords (e.g., crime, violation, fighting), while the normal pool contains 58 keywords (e.g., safety, routine, trustworthy) (A). All keywords are listed in Appendix A and used in upper case. These expanded pools are generated by utilizing GPT-5 [7]. We compute the embeddings for all keywords within each pool and perform score fusion via the difference of means to derive a robust final anomaly score for each frame. The final fused score contrasts the mean similarity of abnormal keywords against the mean similarity of normal keywords.

The intuition is that if an input aligns more strongly with abnormal descriptors than with normal ones, the fused score, will be positive and indicate abnormality. Conversely, if the input is closer to normal descriptors, the score will be negative and indicate normality. This contrastive formulation provides a simple yet effective way to leverage semantic keyword pools for anomaly detection. Note that for instruct models, the text prompts are updated to suppress detailed responses, i.e., *"Reply only with the following word:"* keyword.

### 3.1 Results

We presented result of VLMs that serve as the backbone of AiSearch in Table 1. Various VLMs, achieved AUC scores ranging from 47.46% to 81.54%, with the best result obtained using

**Table 1:** Frame-level AUC performance benchmark on UCF-Crime.

| Model | AUC% |
|---|---|
| GME-Qwen2-VL-2B [13] | 47.46 |
| MMRet-base [14] | 68.14 |
| VLM2Vec (Phi-3.5-V) [3] | 78.22 |
| VLM2Vec (LLaVA-1.6) [3] | 79.85 |
| VLM2Vec (Qwen2-VL-2B) [3] | 69.10 |
| VLM2Vec (Qwen2-VL-7B) [3] | 71.93 |
| VLM2Vec-v2 (Qwen2-VL-2B) [6] | 76.50 |
| Ops-MM-embedding-v1-2B [8] | 77.45 |
| Ops-MM-embedding-v1-7B [8] | **81.54** |
| Qwen3-VL-Embedding-2B [4] | 77.46 |
| Qwen3-VL-Embedding-8B [4] | 78.40 |
| SmolVLM2-256M [5] | 66.68 |
| SmolVLM2-500M [5] | 66.13 |
| Qwen2-VL-2B-Instruct [11] | 68.87 |
| Qwen3-VL-2B-Instruct [1] | 75.26 |
| Qwen3-VL-4B-Instruct [1] | 69.59 |
| Qwen3-VL-8B-Instruct [1] | 74.03 |

Ops-MM-embedding-v1-7B [8]. While Ops-MM-embedding-v1-7B achieves the best score for anomaly detection on the UCF-Crime dataset, other VLMs might perform better in other domains or datasets.

### 3.2 Ablation Study

We conduct an ablation study to evaluate the impact of different prompting strategies and keyword pool sizes on the framework's zero-shot performance, as shown in Table 2. In this analysis, we select six representative models from both the embedding-based and instruct-based categories. We ablated our prompting strategies in three levels:

**Table 2:** Ablation study on prompting strategies and keyword pool sensitivity.

| Model | AUC% | | | | | |
|---|---|---|---|---|---|---|
| | Single Keyword | Multi-Keyword | Pool 25% | Pool 50% | Pool 75% | Pool 100% |
| VLM2Vec (LLaVA-1.6) | 74.7 | 71.6 | 78.9 ± 0.8 | 79.5 ± 0.6 | 79.7 ± 0.5 | 79.9 |
| Ops-MM-emb.-v1-7B | 69.2 | 76.8 | **81.2 ± 0.7** | **81.0 ± 1.2** | **81.7 ± 0.4** | **81.5** |
| Qwen3-VL-Emb.-8B | **77.7** | **78.2** | 78.0 ± 0.6 | 78.2 ± 0.2 | 78.8 ± 0.5 | 78.4 |
| SmolVLM2-256M | 57.6 | 53.7 | 64.4 ± 3.3 | 64.7 ± 1.2 | 65.4 ± 0.3 | 6.7 |
| Qwen2-VL-2B-Inst. | 63.4 | 64.0 | 68.5 ± 0.7 | 68.3 ± 0.5 | 68.3 ± 0.3 | 68.9 |
| Qwen3-VL-2B-Inst. | 72.1 | 62.9 | 69.9 ± 2.8 | 72.0 ± 2.0 | 74.0 ± 0.1 | 75.3 |

**- Single Keyword Prompting -** uses a single keyword for both pools, specifically the generic terms ‘Abnormal’ and ‘Normal’.
**- Multi-Keyword Prompting -** uses the video labels from the UCF-Crime dataset as keywords. This expands the abnormal pool to include the 13 specific crime labels (Abuse, Arrest, Arson, Assault, Burglary, Explosion, Fighting, Road Accident, Robbery, Shooting,

Shoplifting, Stealing, and Vandalism) while maintaining a single 'Normal' keyword, reflecting the single category of videos with normal scenes.

**- Pool Prompting -** utilizes our proposed comprehensive pools of 84 abnormal and 58 normal terms.

To further investigate the sensitivity of the models to pool size, we perform a subset analysis by randomly sampling the pools at ratios of 25%, 50%, and 75%. For each ratio, we execute three independent trials with different random seeds and report the mean and standard deviation of the AUC scores.

From the experimental results in Table 2, we observe that generally Pool Prompting works better than Single Keyword and Multi-Keyword, boosting performance by up to 9.11% (4.53% on average). This indicates that the fused scoring of diverse relevant keywords is more robust than generic or class-specific labels. Qwen3-VL-Embedding-8B [4] achieves the highest scores in Single and Multi-Keyword across the six models, while Ops-MM-embedding-v1-7B [8] outperforms others in Pool Prompting, reaching AUC scores 81.7 $\pm$ 0.41.

### 3.3 Qualitative Results

Figure 2 shows examples of retrieval outputs using our UI design, utilizing (a) text input only; (b) image input only; and (c) a combination of text and image input. Our UI and framework offer high flexibility for users to interact with and explore large-scale video data in real-time, ranging from general queries (e.g. "crime scene", "robbery") to fine-grained, specific searches (e.g., "a woman stealing in a supermarket", "a road accident involving a car and a bus").

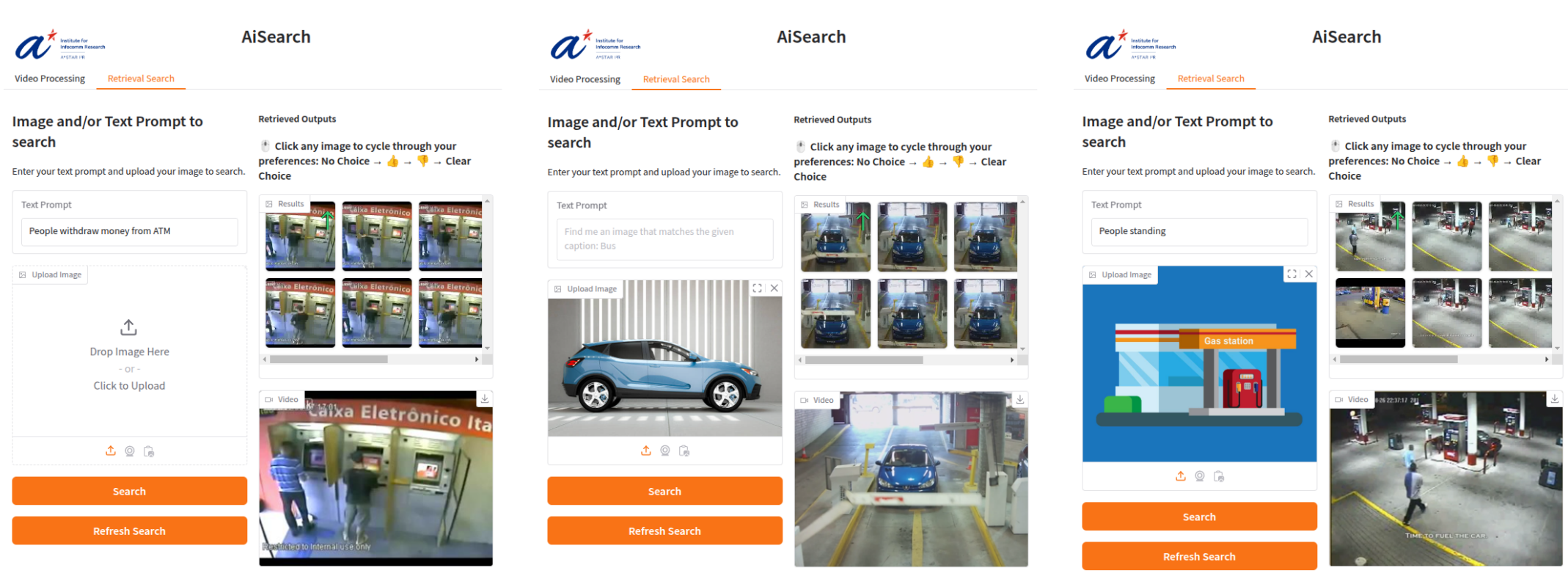

(a) Text Input (b) Image Input (c) Text+Image Input

**Fig. 2:** AiSearch UI design for multi-modal retrieval search. Demo can be viewed in the demo video. Note that since our UI offers flexibility, there might be a slight (positional) difference from what is shown in the figure.

The left-hand side of the UI is used for the input prompt search, and the right-hand side shows the retrieved outputs. On the retrieved outputs, the frame-based and the video output that corresponds to the user's query are shown in the top and the bottom part, respectively. To improve the incident-review workflow, selecting an image triggers the video player to seek to a timestamp corresponding to several frames prior to the chosen image.

## A Keyword Pools for Anomaly Detection

**Abnormal Keywords Pool:** *Abuse, Domestic Violence, Child Abuse, Mistreatment, Arrest, Police Intervention, Detention, Apprehension, Arson, Fire, Blaze, Burning Incident, Assault, Battery, Physical Attack, Stabbing, Burglary, Break-in, Trespassing, Home Invasion, Explosion, Blast, Bombing, Explosion Hazard, Fighting, Brawl, Rioting, Protest Violence, Road Accidents, Traffic Collision, Hit-and-Run, Traffic Violation, Robbery, Mugging, Armed Robbery, Theft with Violence, Shooting, Gunfire, Homicide, Weapon Threat, Shoplifting, Theft, Pickpocketing, Stealing from Store, Stealing, Larceny, Looting, Property Theft, Vandalism, Property Damage, Graffiti, Defacement, Abnormal, Anomaly, Irregular, Unusual, Unexpected, Atypical, Outlier, Deviation, Crime, Criminal Act, Illegal, Unlawful, Felony, Offense, Violation, Infraction, Suspicious, Dangerous, Threatening, Hostile, Violent, Disorderly, Chaotic, Disturbance, Incident, Event, Occurrence, Emergency, Hazard, Disaster, Catastrophe, Crisis*

**Normal Keywords Pool:** *Care, Kindness, Support, Release, Freedom, Acquittal, Safety, Peaceful Interaction, Hug, Handshake, Secure Home, Lawful Entry, Invitation, Calm Environment, Safe Construction, Friendship, Cooperation, Teamwork, Safe Driving, Smooth Traffic, Road Safety, Honest Transaction, Gift Giving, Charity, Disarmament, Peace, Shopping, Purchase, Payment, Sharing, Borrowing with Permission, Ownership, Maintenance, Beautification, Normal, Regular, Routine, Ordinary, Usual, Standard, Commonplace, Typical, Legal, Lawful, Legitimate, Authorized, Permitted, Valid, Licensed, Official, Safe, Secure, Peaceful, Orderly, Disciplined, Controlled, Trustworthy, Respectful*